\documentclass[11pt, a4paper, logo, twocolumn, copyright]{googledeepmind}

\usepackage[authoryear, sort&compress, round]{natbib}
\usepackage[textwidth=18mm]{todonotes}
\usepackage{graphicx}
\usepackage{makecell}
\usepackage{booktabs}
\usepackage{array}
\usepackage{multirow}
\usepackage{tikz}
\usepackage{xcolor}
\usepackage{float}
\usetikzlibrary{shapes.arrows, positioning, shadows, calc}
\usepackage{algorithm}
\usepackage{algpseudocode}
\usepackage{amsmath}
\usepackage{amssymb}
\usepackage{enumitem}
\usepackage{multicol}
\usepackage{xspace}
\usepackage{tabu}
\usepackage{verbatimbox}
\usepackage{xurl}

\definecolor{gdm_darkgray}{HTML}{9aa0a6}
\definecolor{gdm_gray}{HTML}{3c4043}

\newcommand{\ie}{\textit{i}.\textit{e}. }

\renewcommand{\today}{2026-06-19}

\title{Gemma 4 Technical Report}

\author{Gemma Team, Google DeepMind\authfootnotemark{1}}

\authfootnotetext{1}{See Contributions and Acknowledgments section for full author list. Please send correspondence to {\tt gemma4report@gmail.com}.}

\begin{abstract}
We introduce Gemma 4, a new generation of open-weight, natively multimodal language models in the Gemma model family. Designed to advance compute efficiency and reasoning, the Gemma 4 model suite features dense and Mixture-of-Experts architectures, ranging from 2.3B to 31B parameters. Alongside improved vision and audio encoders for all model sizes, we propose a unified, encoder-free architecture for our 12B model, which ingests raw audio and image patches. Furthermore, we integrate a thinking mode, enabling Gemma models to generate reasoning traces prior to responding. We improve inference speed, memory, and compute efficiency, as well as long-context abilities through critical design choices. Gemma 4 establishes a leap in performance across STEM, multimodal, and long-context benchmarks, and rivals larger, frontier open models in human-rated tasks.
\end{abstract}

\begin{document}

\maketitle

\section{Introduction}
The rapid evolution of large language models has driven the need for open-weight models with strong multimodal understanding, reasoning, and computational efficiency. Building upon the foundations of its predecessors~\citep{gemmateam2024gemma,gemmateam2024gemma2,gemmateam2025gemma}, we introduce Gemma 4, the most capable and efficient generation in the Gemma model family to date. Gemma 4 offers natively multimodal architectures, capable of seamlessly processing text, images, and audio while achieving frontier-level performance on highly complex reasoning tasks.
The Gemma 4 family is built to serve a variety of on-device hardware. The model suite includes both dense architectures (2.3B, 4.5B, 12B, and 31B parameters) and a Mixture-of-Experts~\citep[MoE]{jacobs1991adaptive} variant with 3.8B activated and 26B total parameters.
We introduce several architectural and methodological innovations:
\begin{itemize}[itemsep=1pt,leftmargin=10pt]
\item \textbf{Thinking mode for advanced reasoning:} We introduce a thinking mode~\citep{jaech2024openai} to Gemma 4 models. By outputting a reasoning trace before the response, models demonstrate improved capabilities in reasoning-heavy domains such as mathematics and coding.
\item \textbf{Long-context efficiency:} Extended contexts lead to a memory explosion in the KV cache. We conserve a 5:1 ratio of local sliding window to global self-attention (4:1 for the 2.3B model) and use $p$-RoPE~\citep{barbero2025round} as positional encoding. Combined with KV cache sharing~\citep{mqa} and the reuse of keys as values in global layers~\citep{kayyam2026transformers}, these optimizations reduce the global KV cache footprint by up to 37.5\%.
\item \textbf{Compute efficiency:} We release an autoregressive multi-token prediction (MTP) drafter head~\citep{li2024eagle} designed for speculative decoding~\citep{10.5555/3618408.3619203} to improve the decoding speed of our models. 
\item \textbf{Memory efficiency:} We provide quantized versions of our models trained with quantization-aware training~\citep[QAT]{jacob2018quantization} to reduce their parameter memory footprint and latency with minimal impact on quality.
\item \textbf{Encoder-free architecture}: Gemma 4 models have frozen vision and audio encoders. We introduce a unified encoder-free architecture for the 12B model, which projects raw 40ms audio chunks and image patches into the LLM embedding space, alleviating the need for separate encoders and reducing memory fragmentation.
\end{itemize}

In this technical report, we outline the different model architectures across model sizes as well as the pre-training and post-training recipe of Gemma 4. Through comprehensive benchmarks and human evaluations such as Arena~\citep{chiang2024chatbot}, we demonstrate that Gemma 4 operates at a level comparable to larger, frontier open-source models across text, image, and audio modalities. We release the Gemma 4 models under an Apache 2.0 license, empowering developers and researchers everywhere to build upon, customize, and extend these capabilities.

\begin{table*}[t!]
    \centering
    \begin{tabular}{@{}l r r r r r@{}}
    \toprule
        Model & \makecell{Audio\\ Encoder} & \makecell{Vision\\ Encoder} & \makecell{Embedder} & \makecell{Einsums} & \makecell{Drafter} \\
        \midrule
        \textbf{E2B}  & 305M & 150M & \makecell{400M + 2,340M} & 1,870M & 76M \\
        \textbf{E4B}  & 305M & 150M & \makecell{670M + 2,820M} & 3,940M & 77M \\
        \textbf{12B} & - & - & 1,000M & 10,890M & 400M \\
        \textbf{26B-A4B*} & - & 550M & 740M & 24,500M  / 2,800M (active) & 430M \\
        \textbf{31B} & - & 550M & 1,410M & 29,290M & 500M \\
    \bottomrule
    \end{tabular}
    \caption{Parameter counts for the Gemma~4 models. 
    The vocabulary we use has 262k entries. The model noted with a star is an MoE defined by its number of active parameters. Note that the extra embedder parameters in E2B and E4B are per-layer embeddings.
    }
    \label{tab:model_param_counts}
\end{table*}

\section{Model Architecture}
\label{sec:architecture}

Gemma 4 models follow a decoder-only Transformer architecture~\citep{DBLP:journals/corr/VaswaniSPUJGKP17}. Our models have pre-norm and post-norm with~RMSNorm~\citep{rmsnorm}, and~QKNorm~\citep{henry2020query}. 

\noindent\textbf{Dense and MoE:} The Gemma 4 family of models comprises dense architectures, with effective~2.3B~(\textbf{E2B}), effective 4.5B (\textbf{E4B}), \textbf{12B} and~\textbf{31B} parameters, as well as an MoE model with 3.8B activated parameters for 26B total parameters (\textbf{26B-A4B}).
E2B and E4B use per-layer embeddings as in Gemma 3n~\citep{google2025gemma3n}, making them 2.3B and 4.5B effective out of 5B and 8B total parameters respectively.

\begin{table}[h!]
 \setlength{\tabcolsep}{5pt}
    \centering
    \begin{tabular}{@{}l c c c c c@{}}
    \toprule
    & & & \multicolumn{3}{c}{Shards} \\
    \cmidrule{4-6}
    Model & TPU & \#Chips & Data & Seq & Replica \\
    \midrule
        \textbf{E2B} & v6e & 4,096 & 16 & 8 & 32 \\
        \textbf{E4B} & v6e & 6,144 & 16 & 16 & 24 \\
        \textbf{12B} & v4 & 12,288 & 16 & 16 & 48 \\
        \textbf{26B-A4B*} & v6e & 6,144 & 16 & 16 & 24 \\
        \textbf{31B} & v6e & 10,240 & 16 & 16 & 40 \\
    \bottomrule
    \end{tabular}
    \caption{Pre-training infrastructure with sharding by data, sequence (\textit{Seq}), and replica.}
    \label{tab:training_infra_sharding}
\end{table}

\noindent\textbf{Long-context efficiency:} Our local to global attention ratio patterns follow~\cite{gemmateam2025gemma}, that is, 4-to-1 local attention blocks for~E2B and 5-to-1 for the rest. We improve memory efficiency by re-using keys as values in the global attention layers (except in~E2B and~E4B),~\ie, $\text{values} = \text{keys}$. We encode position with $p$-RoPE with $p=0.25$ on global attention layers and with RoPE on local attention layers, effectively reducing the global KV cache by 37.5\%. The RoPE frequencies are set to 1M and 10k on global and local attention layers, respectively. Finally, we share the KV cache with ratios of~20/35 and~18/42 for the E2B and E4B model.

\vspace{-1em}
\subsection{Vision modality}

E2B and E4B Gemma models come with a 150M vision encoder, while larger models use a~550M encoder (except for the unified 12B). Both are Vision Transformers~\citep[ViT]{dosovitskiy2020image} with a patch size of 16, whose architectural differences are detailed in Table \ref{tab:vision_encoder_comparison} in Appendix.
Our vision encoders support variable aspect ratios~(see Figure~\ref{fig:varres} and Algorithm~\ref{alg:resize}) and incorporate both axial 2D-RoPE~\citep{heo2024rotary} with non-causal attention and 2D absolute positional embeddings. We restrict the maximum number of tokens, $N_\text{max}$ to the values~$70,140, 280, 560$ and~$1120$~(see Algorithm~\ref{alg:resize} for implementation details).

\subsection{Audio modality}
E2B and E4B Gemma models use a 305M audio encoder that processes audio in 40ms chunks with Mel filterbank inputs. The encoder architecture is based on the Universal Speech Model~\citep[USM]{zhang2023google}, consisting of two downsampling convolution layers followed by twelve Conformer layers~\citep{gulati2020conformer}. While the architecture remains similar to that of Gemma~3n, we reduce the number of parameters by~55\%~(from 680M to 305M). We do not use vector quantization; the LLM ingests the continuous representations produced by the audio encoder. As with the vision encoder, we keep weights frozen during pre-training.

\subsection{Encoder-free architecture}
Gemma 4 12B is trained from scratch based on a new, unified, and encoder-free model paradigm, replacing the separate vision and audio encoders with lightweight projection modules. For the vision modality, Gemma 4 12B takes in~48$\times$48$\times$3~RGB patches, but replaces the 550M vision encoder by a single large matmul (35M parameters). Spatial awareness is maintained by adding 2D coordinate-based positional embeddings directly to the patch representations before a final LayerNorm layer~\citep{ba2016layer}.

\noindent{} For audio, the 305M USM-based conformer encoder is \textit{entirely discarded}. Raw audio is segmented into 40ms chunks at 16kHz, resulting in 640-dimensional vectors per chunk. These are projected directly into the LLM embedding space. Since audio is a temporal sequence, it does not require additional positional encoding.

\begin{table}[h!]
 \setlength{\tabcolsep}{5pt}
    \centering
    \begin{tabular}{@{}l c c c@{}}
    \toprule
    Model & bf16 & Quantized & KV Cache \\
    \midrule
        \textbf{E2B} & 4.6 & ${0.8}^{\dagger}$ & +0.05 \\
        \textbf{E4B} & 9.0 & ${2.3}^{\dagger}$ & +0.14 \\
        \textbf{12B} & 24.0 & ${7.65}^{\ddagger}$ & +0.28 \\
        \textbf{26B-A4B*} & 52.0 / 7.6 & ${16.2 / 2.8}^{\ddagger}$ & +0.28 \\
        \textbf{31B} & 64.0 & ${19.2}^{\ddagger}$ & +1.10 \\
    \bottomrule
    \end{tabular}
    \caption{Text only, Gb memory footprint comparison between raw and quantized checkpoints for weights and int8 KV caching (+KV) at 32k context size. $\dagger$ is mobile quantization, $\ddagger$ is Q4\_0.}
    \label{tab:quantization}
\end{table}

\subsection{Pre-training}

We follow a similar pre-training as Gemma~3.

\noindent\textbf{Training data.} Our pre-training dataset is a large-scale, diverse collection of data from a wide range of domains and modalities, including web documents, code, images, and audio (for E2B, E4B and 12B), with a cutoff date of January 2025.

\begin{figure}[t]
    \centering
    \includegraphics[width=\linewidth]{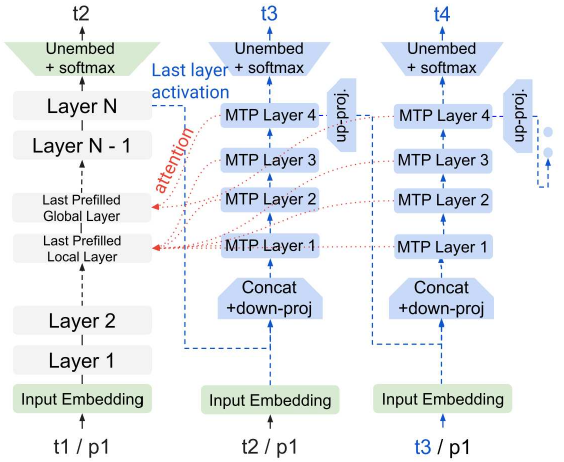}
    \caption{The autoregressive MTP drafter (blue blocks on the right) is fed activations and KV cache from the main model (gray blocks).}
    \label{fig:mtp}
\end{figure}

\noindent\textbf{Tokenizer.}
We use the same tokenizer as~\citet{comanici2025gemini} that is, a SentencePiece tokenizer~\citep{kudo-richardson-2018-sentencepiece} with split digits, preserved whitespace, and byte-level encodings.
The vocabulary has 262k entries.

\noindent\textbf{Filtering.} We  filter data to decontaminate benchmarks, and to reduce the risk of unwanted or unsafe utterances and the risk of recitation.

\begin{table*}[t]
\centering
\begin{tabular}{@{}llccccc@{}}
\toprule
Rank & Model & Elo   & 95\% CI        & Open       &Type & \#params/\#activated\\
\midrule
\color{gray} 1       & \color{gray} Claude Fable 5       & \color{gray} 1508  & \color{gray} $\pm$ 9 & \color{gray} no & \color{gray} - & \color{gray} - / - \\
...     &       &       &       &       &       & \\
15      & GLM 5.1 \nocite{zeng2026glm}       & 1475  & $\pm$ 6 & yes & MoE & 744B / 40B\\
25      & GLM 5.2 (Max) & 1471  & $\pm$ 10 & yes & MoE & 744B / 40B\\
29      & MiMo V2.5 Pro \nocite{xiao2026mimo} & 1466  & $\pm$ 5 & yes & MoE & 1T / 42B\\
34      & Kimi K2.6 \nocite{team2026kimi}    & 1460  & $\pm$ 5 & yes & MoE & 1T / 32B\\
36      & DeepSeek V4 Pro Thinking \nocite{xu2026deepseek}  & 1458  & $\pm$ 5 & yes & MoE & 1.6T / 49B\\
37      & GLM 5         & 1457  & $\pm$ 5 & yes & MoE & 744B / 40B\\
38      & DeepSeek V4 Pro  & 1456  & $\pm$ 5 & yes & MoE & 1.6T / 49B\\
\color{blue!70!black} 43      & \color{blue!70!black} \textbf{Gemma 4 31B}        & \color{blue!70!black}\textbf{1451}  & \color{blue!70!black} \textbf{$\pm$ 8} & \color{blue!70!black} \textbf{yes} & \color{blue!70!black} \textbf{Dense} & \color{blue!70!black} \textbf{31B}\\
44      & Kimi K2.5 Thinking & 1450  & $\pm$ 4 & yes & MoE & 1T / 32B\\
57      & Qwen 3.5 397B-A17B \nocite{team2026qwen3} & 1444  & $\pm$ 4 & yes & MoE & 397B / 17B\\
\color{blue!70!black} 61      & \color{blue!70!black} \textbf{Gemma 4 26B-A4B}        & \color{blue!70!black}\textbf{1438}  & \color{blue!70!black} \textbf{$\pm$ 8} & \color{blue!70!black} \textbf{yes} & \color{blue!70!black} \textbf{MoE} & \color{blue!70!black} \textbf{26B / 4B}\\
63      & DeepSeek V4 Flash Thinking~~         & 1436  & $\pm$ 5 & yes & MoE & 284B / 13B\\
...     &       &       &       &       &       & \\
157      & Gemma 3 27B         & 1366  & $\pm$ 4 & yes & Dense & 27B\\
\bottomrule
\end{tabular}
\caption{Leading open-weight models on Arena Text~\citep{chiang2024chatbot} (as of June 19, 2026). Models are evaluated through blind side-by-side evaluations by human raters, and attributed scores based on the Elo rating system. The top closed model (gray) is included for scale. Gemma models rival much larger models, and Gemma 4 31B is the leading dense open model on the leaderboard.}
 \label{tab:lmsys_elo_leaderboard}
\end{table*}

\subsection{Quantization-Aware Training}

We provide quantized models and encoders in different formats along with the raw checkpoints. 
Based on the most popular open source quantization inference engines (e.g. \texttt{llama.cpp}) as well as efficient hardware support, we focus on two sets of weight representations:
\begin{itemize}[itemsep=10pt,leftmargin=10pt,topsep=9pt]
    \item mobile quantization: per-channel low bitwidth weight (mix of int2 and int4) and activation quantization (int8).
    \item Q4\_0 quantization: blockwise quantization, often referred to as Q4\_0.
\end{itemize}
In Table~\ref{tab:quantization}, we report the memory filled by raw and quantized models with and without a KV cache for a sequence of 32k tokens.
Furthermore, to enable stable inference in fp16, we introduce a scalar scale at each block in order to bound the activation ranges to fit fp16.

\begin{table*}[th!]
\centering
\begin{tabular}{@{}l ccccc c c@{}}
\toprule
& \multicolumn{5}{c}{Gemma~4} && Gemma~3 \\
\cmidrule{2-6}\cmidrule{8-8}
& 31B & 26B-A4B & 12B & E4B & E2B && \makecell{27B\\{\tiny non-thinking}} \\
\midrule
MMLU Pro \nocite{wang2024mmlu} & 85.2 & 82.6 & 77.2 & 69.4 & 60.0 && 67.6 \\
\midrule
AIME 2026 {\tiny no tools} & 89.2 & 88.3 & 77.5 & 42.5 & 37.5 && 20.8 \\
\midrule
LiveCodeBench v6 \nocite{jain2025livecodebench} & 80.0 & 77.1 & 72.0 & 52.0 & 44.0 && 29.1 \\
Codeforces {\tiny Elo} & 2150 & 1718 & 1659 & 940 & 633 && 110 \\
SciCode \nocite{tian2024scicode} & 43.0 & 40.0 & 38.0 & 24.0 & 21.0 && 21.0 \\
\midrule
GPQA Diamond \nocite{Rein2023GPQAAG} & 84.3 & 82.3 & 78.8 & 58.6 & 43.4 && 42.4 \\
\midrule
Big Bench Extra Hard {\tiny micro avg}\nocite{kazemi2025big} & 74.4 & 64.8 & 53.0 & 33.1 & 21.9 && 19.3 \\
\midrule
HLE \nocite{phan2025humanity} & 19.5 & 8.7 & 5.2 & - & - && - \\
HLE with search \nocite{center2026benchmark} & 26.5 & 17.2 & - & - & - && - \\
\midrule
IFBench \nocite{pyatkin2026generalizing} & 76.0 & 72.0 & 74.0 & 44.0 & 38.0 && 32.0 \\
IFEval \nocite{zhou2023instruction} & 98.9 & 98.5 & 97.2 & 96.7 & 94.6 && 90.4 \\
\midrule
MMMLU & 88.4 & 86.3 & 83.4 & 76.6 & 67.4 && 70.7 \\
\midrule
MRCR v2 {\tiny 8-needle, 128k} \nocite{vodrahalli2024michelangelo} & 66.4 & 44.1 & 43.4 & 25.4 & 19.1 && 13.5 \\
\midrule
Terminal Bench Hard \nocite{merrill2026terminal} & 36.0 & 14.0 & 18.0 & 8.0 & 3.0 && 4.0 \\
Tau2 -- airline \nocite{barres2025tau2} & 75.0 & 76.0 & 75.0 & 52.0 & 31.0 && 39.0 \\
Tau2 -- retail & 86.4 & 85.5 & 77.6 & 67.1 & 34.6 && 6.6 \\
Tau2 -- telecom & 69.3 & 43.0 & 54.4 & 18.4 & 19.7 && 3.1 \\
\bottomrule
\end{tabular}
\caption{Performance comparison of Gemma~3 27B and Gemma~4 models on diverse benchmarks. All models are in thinking mode unless explicitly stated.}
\label{tab:it_fs}
\end{table*}

\noindent{}We also apply QAT to the image and audio encoders. On the 150M image encoder, quantizing activations and weights to 8-bit precision~(W8A8) yields a 2$\times$ reduction in total forward-pass memory footprint (from 400 MB to 200 MB, including on-device compilation overhead) and a 44\% reduction in on-device latency relative to Gemma~3n on newer hardware.
On the audio encoder, we further reduce activation precision to 8 bits and weight precision to $\{2, 4, 8\}$ bits, varying by layer cluster. Overall, we achieve a 78\% reduction in on-disk footprint, from 390\,MB in Gemma 3n to~87\,MB in this version.

\subsection{Multi-Token Prediction Drafter}

We train a small autoregressive MTP drafter head with our models, used for speculative decoding. In our MTP procedure, the model's last layer activations from the previous step and token embeddings are fed into the MTP head. The MTP head generates future tokens sequentially using a separate embedder and a 4-layer Transformer block that cross-attends to the KVs of the main model (Figure~\ref{fig:mtp}), thus eliminating the need for MTP prefill and supporting any draft length. The Transformer block has model dimension 256 for E2B and E4B, 1024 for 26B-A4B and 31B, three local, and one global attention layers.

\paragraph{Efficient MTP Decoding.}
For the E2B and E4B drafters, we reduce the decoding overhead by replacing the projection operation to the entire vocabulary by a top-k operation on clusters of tokens. As a result, final matrix multiplication is reduced from $d\times 262,000$ to $d\times 4096$ while preserving a similar acceptance rate.

\subsection{Compute Infrastructure}

We train our models with TPUv4 and TPUv6e as outlined in Table~\ref{tab:training_infra_sharding}.
Each model configuration is optimized to minimize training step time. 
For our larger models, we leverage Slice-Granularity Elasticity~\citep{comanici2025gemini}, which allows continuous training with fewer “slices” of TPU chips when there is a localized failure. This reconfiguration reduces the delay caused by interruptions from many minutes to a few seconds.

The optimizer state is sharded using an implementation of ZeRO-3~\citep{ren2021zero}. 
For multi-pod training, we perform a data replica reduction over the data center network, using the Pathways approach of \citet{barham2022pathways}.
We use the single controller programming paradigm of JAX~\citep{bradburyJAX} and Pathways, along with the GSPMD partitioner~\citep{gspmd} and the MegaScale XLA compiler~\citep{xla}.

\begin{table*}[th!]
\centering
\setlength{\tabcolsep}{7pt} 
\begin{tabular}{@{}l ccccc c c @{}}
\toprule
 & \multicolumn{5}{c}{Gemma~4} && Gemma~3 \\
\cmidrule{2-6}\cmidrule{8-8}
& 31B & \makecell{26B-A4B} & 12B & E4B & E2B && 27B \\
\midrule
MMMU Pro \nocite{yue2025mmmu} & 76.9  & 73.8  & 69.1  & 52.6 & 44.2 && 49.7 \\
MATH-Vision \nocite{wang2024measuring} & 85.6 & 82.4  & 79.7  & 59.5  & 52.4 && 46.0 \\
MedXPertQA MM \nocite{zuo2025medxpertqa} & 61.3 & 58.1 & 48.7 & 28.7 & 23.5  && - \\
InfographicVQA \nocite{mathew2022infographicvqa} & 92.0  & 89.3  & 88.4  & 70.0 & 63.9 && 70.6 \\
OmniDocBench 1.5 $\downarrow$ \nocite{ouyang2025omnidocbench} & 0.131 & 0.149 & 0.164 & 0.181  & 0.290 && 0.365 \\
\bottomrule
\end{tabular}
\caption{Gemma~4 models performance on vision benchmarks at different resolutions (thinking). We use the maximal supported resolution (1120 vision tokens) and report results with 280 vision tokens in Table~\ref{tab:it_vis_sl_280}. Gemma~3 27B is non-thinking and uses Pan \& Scan.}
\label{tab:it_vis_sl}
\vspace{3em}
\end{table*}

\section{Instruction Tuning}
Pre-trained models are turned into instruction-tuned models with a similar post-training approach as in Gemma 3. A significant difference is the addition of a thinking mode, where the model can output a reasoning trace before answering.

\noindent\textbf{Data filtering.} We carefully optimize the data used in post-training to maximize model performance. We filter examples that show certain personal information, unsafe or toxic model outputs, mistaken self-identification data, and duplicated examples. Including subsets of data that encourage better in-context attribution, hedging, and refusals to minimize hallucinations also improves performance on factuality metrics, without degrading model performance on other metrics. 

\noindent\textbf{PT versus IT formatting.} 
All models share the same tokenizer, with some control tokens dedicated to IT formatting.
A key difference is that PT models output an \texttt{<eos>} token at the end of generation, while IT models output \texttt{<turn|>} at the end of the generation. An example is given for IT in Table~\ref{tab:formatting_tokens}.
Fine-tuning either model type thus requires adding their respective end tokens. We detail how to activate thinking and how models handle function calling in Table~\ref{tab:formatting_tokens}.

\phantomsection
\section{Evaluation of final models}
\label{sec:evals}

In this section, we evaluate the IT models over a series of automated benchmarks and human evaluations across a variety of domains, as well as static benchmarks such as MMLU Pro.

\subsection{Human evaluation}

We report the performance of our 31B and 26B-A4B models on Arena \citep{chiang2024chatbot} in blind side-by-side evaluations by human raters against other state-of-the-art models. 
We report Elo scores in Table \ref{tab:lmsys_elo_leaderboard}. 
Gemma 4 31B is the top open model in the dense category, and both Gemma 4 31B and 26B-A4B show performance equal to much larger open models.

\begin{table*}[th!]
\footnotesize
\centering

\setlength{\tabcolsep}{4pt}
\resizebox{\textwidth}{!}{%
\begin{tabular}{@{}l cc *{8}{c} @{}}
\toprule
\multicolumn{11}{l}{\textit{CoVoST \nocite{wang2020covost} (CorpusBLEU $\uparrow$ )}} \\
\midrule
 & Params & Size & ja $\to$ en & de $\to$ en & fr $\to$ en & es $\to$ en & it $\to$ en & ru $\to$ en & zh $\to$ en & AVG \\
\midrule
Gemma 4 E2B    & \multirow{2}{*}{305M} & \multirow{2}{*}{87\,MB} & 21.4 & 39.2 & 39.2 & 43.2 & 40.8 & 46.4 & 17.9 & 35.4 \\
Gemma 4 E4B    &  &  & 25.5 & 42.0 & 41.0 & 44.8 & 43.0 & 49.4 & 21.9 & 38.2 \\
\midrule
Gemma 3n E2B    & \multirow{2}{*}{680M} & \multirow{2}{*}{390\,MB} & 17.7 & 36.5 & 35.7 & 39.9 & 38.5 & 39.2 & 13.9 & 31.6 \\
Gemma 3n E4B    &  &  & 22.3 & 39.1 & 38.4 & 41.8 & 40.4 & 43.7 & 17.4 & 34.7 \\
\bottomrule
\end{tabular}%
}

\vspace{1.5em}

\resizebox{\textwidth}{!}{%
\begin{tabular}{@{}l *{13}{c} @{}}
\toprule
\multicolumn{14}{l}{\textit{FLEURS ASR \nocite{conneau2023fleurs} (WER $\downarrow$\, , \textsuperscript{*} = CER $\downarrow$ )}} \\
\midrule
 & en & ko\textsuperscript{*} & ja\textsuperscript{*} & de & fr & hi & es & it & pt-br & ru & ar & zh\textsuperscript{*} & AVG \\
\midrule
Gemma 4 E2B    & 0.080 & 0.066 & 0.107 & 0.076 & 0.101 & 0.101 & 0.042 & 0.041 & 0.056 & 0.084 & 0.143 & 0.187 & 0.090 \\
Gemma 4 E4B    & 0.065 & 0.053 & 0.078 & 0.061 & 0.080 & 0.086 & 0.035 & 0.032 & 0.046 & 0.068 & 0.162 & 0.136 & 0.075 \\
\midrule
Gemma 3n E2B    & 0.076 & 0.101 & 0.163 & 0.079 & 0.130 & 0.106 & 0.051 & 0.044 & 0.067 & 0.112 & 0.131 & 0.235 & 0.108 \\
Gemma 3n E4B    & 0.066 & 0.073 & 0.111 & 0.065 & 0.098 & 0.089 & 0.041 & 0.034 & 0.053 & 0.087 & 0.101 & 0.203 & 0.085 \\
\bottomrule
\end{tabular}%
}

\caption{Audio performance for Gemma~4 and Gemma~3n models. Top: CoVoST (S2TT prompt: \textit{transcribe then translate}). Bottom: FLEURS ASR (transcription). Compared to Gemma~3n of corresponding sizes, Gemma~4 achieves a 12\% (E2B) / 10\% (E4B) relative improvement on translation and a 17\% (E2B) / 12\% (E4B) relative improvement on transcription, despite a 78\% reduction in on-disk audio encoder footprint (from 390\,MB to 87\,MB after quantization).}
\label{tab:audio_evals}
\end{table*}

\subsection{Static benchmarks}

In Table~\ref{tab:it_fs}, we show the performance of our final models across a variety of benchmarks compared to Gemma 3 27B. Gemma 4 31B is closest in size and significantly better across the board, while E2B roughly matches Gemma 3 27B performance with 10x less parameters.
Table~\ref{tab:it_vis_sl} shows the performance of Gemma 4 models on vision benchmarks, with E4B equaling or outperforming Gemma 3 27B on all evals.
Tables~\ref{tab:audio_evals} and \ref{tab:audio_evals_12b} display the multilingual audio transcription and translation performance of E2B \& E4B and of 12B respectively.
Table~\ref{tab:it_lc} shows a leap on long-context capabilities between Gemma 3 27B and Gemma 4 models, with E4B outperforming Gemma 3 27B.

\section{Responsibility, Safety, Security}
\label{safety}

As open models become central to enterprise infrastructure, provenance and security are paramount. Gemma 4 undergoes the same rigorous safety evaluations as Gemini models. Responsibility, safety, and security are of utmost importance in the development workflow, ensuring that these language models are designed from the ground up for responsible AI development.

\looseness=-1
\subsection{Governance \& Assessment}

Our approach to assessing the benefits and risks of Gemma 4 reflects the foundation established in prior models, updated to account for its expanded multimodal capabilities. We maintain the belief that openness in AI can spread the benefits of these technologies across society, but this must be continuously evaluated against the risk of malicious uses that can cause individual and institutional harm~\citep{weidinger2021ethicalsocialrisksharm}. 

Gemma 4 models were developed in partnership with internal safety and responsible AI teams. Releasing these models required careful scrutiny of the evolving risks associated with LLMs and an understanding of how models are deployed in the wild. While an open model shares innovation across the AI ecosystem, we remain committed to providing educational resources to users and monitoring downstream model usage.

\begin{table}[t!]
\footnotesize
\centering
\setlength{\tabcolsep}{7pt}
\begin{tabular}{@{} *{5}{c} @{}}
\toprule
\multicolumn{5}{@{}c}{\textit{FLEURS ASR (WER $\downarrow$\, , \textsuperscript{*} = CER $\downarrow$ )}} \\
\midrule
en & ko\textsuperscript{*} & ja\textsuperscript{*} & de & fr \\
\midrule
0.063 & 0.057 & 0.080 & 0.053 & 0.081  \\
\midrule
es & it & pt-br & ru & ar \\
\midrule
0.038 & 0.030 & 0.047 & 0.068 & 0.070 \\
\bottomrule
\end{tabular}

\vspace{1.5em}

\begin{tabular}{@{} *{6}{c} @{}}
\toprule
\multicolumn{6}{c}{\textit{CoVoST (XX $\to$ EN, CorpusBLEU $\uparrow$ )}} \\
\midrule
ja & de & fr & es & it & ru \\
\midrule
26.4 & 41.9 & 42.5 & 44.6 & 43.3 & 50.5 \\
\bottomrule
\end{tabular}
\caption{Audio performance of Gemma~4 12B model on supported languages, demonstrating that competitive audio-text performance can be achieved without a dedicated audio encoder.}
\label{tab:audio_evals_12b}
\end{table}

\begin{table*}[th!]
\setlength{\tabcolsep}{5.5pt}
\begin{tabular}{@{}lll ccccc c c@{}}
\toprule
& & & \multicolumn{5}{c}{Gemma~4} && Gemma~3 \\
\cmidrule{4-8}\cmidrule{10-10}
Benchmark & Metric & Context length & 31B & 26B-A4B & 12B & E4B & E2B && 27B \\
\midrule
\multirow{2}{*}{RULER \nocite{hsieh2024ruler}} & \multirow{2}{*}{Accuracy} & 32k & 96.8 & 97.3 & 96.4 & 95.2 & 83.0 && 91.1 \\
 & & 128k & 96.4 & 89.8 & 91.2 & 86.6 & 70.4 && 66.0 \\
\midrule
\begin{tabular}[c]{@{}l@{}}LOFT \\ Text Retrieval\end{tabular} \nocite{Lee2024LongContext} & Recall@k & 128k & 79.5 & 66.3 & 66.4 & 58.5 & 50.5 && 8.6 \\
\midrule
GraphWalks \nocite{openai2025graphwalks} & F1 & <128k & 82.3 & 72.6 & 71.0 & 50.9 & 4.1 && 32.8 \\
\midrule
MTOB  & \multirow{2}{*}{chrF} \nocite{tanzer2024a} & $\sim$128k (Half book) & 52.9 & 50.0 & 45.1 & 37.8 & 15.4 && 41.0\\
~~(eng$\rightarrow$kgv) & & $\sim$256k (Full book) & 54.3 & 48.9 & 41.9 & - & - && - \\
 \midrule
MTOB & \multirow{2}{*}{chrF} & $\sim$128k (Half book) & 48.6 & 45.0 & 37.3 & 34.6 & 28.2 && 31.2 \\
~~(kgv$\rightarrow$eng)& &$\sim$256k (Full book) & 46.2 & 42.7 & 32.9 & - & - && - \\
\bottomrule
\end{tabular}
\caption{Long context performance of Gemma 3 and Gemma 4 models (without thinking).}
\label{tab:it_lc}
\end{table*}

\subsection{Safety Policies and Train-Time Mitigations}

A key pillar of Gemma's safety approach is aligning our fine-tuned models with Google's AI principles and safety policies. These policies aim to prevent our generative models from producing harmful content, specifically:

\begin{itemize}[nosep,leftmargin=10pt,topsep=6pt]
    \item Content related to child sexual abuse material (CSAM) and exploitation;
    \item Dangerous content, e.g., promoting suicide, or instructing in activities that could cause real-world harm;
    \item Sexually explicit content;
    \item Hate speech, e.g., dehumanizing members of protected groups;
    \item Harassment, e.g., encouraging violence against people.
\end{itemize}

To mitigate these risks, Gemma 4 models underwent careful input data pre-processing and scrutiny. The training data was specifically filtered for the removal of certain personal information and other sensitive data to guard against privacy violations. Post-training evaluations and train-time mitigations were also implemented to align the model with our safety policies.

\subsection{Safety Evaluations}

We conduct rigorous automated and human evaluations to understand the potential harms our models might cause. For all areas of safety testing, we saw major improvements in every category of content safety relative to previous Gemma models. Overall, Gemma 4 models significantly outperform Gemma 3 and 3n models in improving safety, while keeping unjustified refusals low. 

Importantly, all testing was conducted without safety filters to accurately evaluate the model's inherent capabilities and behaviors. For both text-to-text and image-to-text modalities, and across all model sizes, the models produced minimal policy violations. We balance development speed with targeted safety testing, upholding the commitments laid out in our Frontier Safety Framework~\citep{DeepMind_2025_strengthening_frontier_safety_framework}.

\subsection{Ethical Considerations and Risk Mitigation}

The development of LLMs introduces specific ethical considerations. In making Gemma 4, we focused heavily on:
\begin{itemize}[itemsep=8pt,leftmargin=10pt]
    \item \textbf{Bias and Fairness}: LLMs trained on large-scale text and image data can reflect embedded socio-cultural biases. We encourage developers to perform continuous monitoring (using evaluation metrics and human review) and explore de-biasing techniques during model fine-tuning.
    \item \textbf{Misinformation and Misuse}: LLMs can be misused to generate false or misleading text. We provide technical limitations, developer education, and guidelines for responsible use within the Responsible Generative AI Toolkit to mitigate malicious applications.
    \item \textbf{Privacy Considerations}: While our training datasets were filtered to remove certain personal information and other sensitive data, developers are strongly encouraged to adhere to local privacy regulations and implement privacy-preserving techniques in their applications.
\end{itemize}

\vspace{-20pt}
\subsection{Our Approach to Responsible Open Models}

Designing safe, secure, and responsible applications requires a system-level approach that mitigates risks associated with specific use cases and environments. We provide guidelines, mechanisms, and safeguards for content safety, and encourage developers to implement appropriate configurations based on their product policies. We will continue to adopt safety mitigations proportionate to potential risks, sharing these models with the community only when confident that the benefits significantly outweigh foreseeable risks.

\section{Discussion and Conclusion}

In this technical report, we presented Gemma 4, an open-weight model family featuring multimodal dense and MoE architectures designed for varied hardware environments. Gemma 4 models come with a thinking mode in which they generate reasoning traces prior to responding, improving overall performance. We introduced a unified, encoder-free architecture that processes raw audio and image patches. We also alleviated long-context memory limitations via better local-to-global attention ratios, positional encoding, and KV cache sharing. We increased the overall compute efficiency via QAT and memory efficiency via MTP drafters. Gemma 4 models demonstrate a leap in performance compared to Gemma 3 across benchmarks, and human evaluations demonstrate that Gemma 4 performs comparably to significantly larger open models, providing a scalable foundation for edge deployment and reasoning while supporting open research.

\bibliography{main}

\onecolumn
\noindent\textbf{Core contributors} \\

\begin{multicols}{3}
\noindent
Sherif El Abd \\ 
Vaibhav Aggarwal \\ 
Robin Algayres \\ 
Alek Andreev \\ 
Olivier Bachem \\ 
Ian Ballantyne \\ 
Cormac Brick \\ 
Victor Cărbune \\ 
Michelle Casbon \\ 
Mayank Chaturvedi \\
Aditya Chawla \\ 
Victor Cotruta \\ 
Alice Coucke \\ 
Phil Culliton \\ 
Robert Dadashi \\ 
Lucas Dixon \\ 
Mohamed Elhawaty \\ 
Utku Evci \\ 
Clément Farabet \\ 
Johan Ferret \\ 
Filippo Galgani \\ 
Sertan Girgin \\ 
Jean-Bastien Grill \\ 
Maarten Grootendorst \\ 
Jiaxian Guo \\ 
Cassidy Hardin \\ 
Yanzhang He \\ 
Steven M. Hernandez \\ 
Omri Homburger \\ 
Léonard Hussenot \\ 
Juyeong Ji \\ 
Armand Joulin \\ 
Aishwarya Kamath \\ 
Parnian Kassraie \\ 
Olivier Lacombe \\ 
Preethi Lahoti \\ 
Gaël Liu \\ 
Gus Martins \\ 
Luciano Martins \\ 
Tatiana Matejovicova \\ 
Ramona Merhej \\ 
Nikola Momchev \\ 
Sneha Mondal \\ 
Ryan Mullins \\ 
Sindhu Raghuram Panyam \\ 
Shreya Pathak \\ 
Sarah Perrin \\
André Susano Pinto \\ 
Etienne Pot \\ 
Angéline Pouget \\ 
Alexandre Ramé \\ 
Sabela Ramos \\ 
Douglas Reid \\ 
David Rim \\ 
Morgane Rivière \\ 
Karsten Roth \\ 
Louis Rouillard \\
Omar Sanseviero \\
Pier Giuseppe Sessa \\ 
Shane Settle \\ 
Danila Sinopalnikov \\ 
Sara Smoot \\ 
Piotr Stanczyk \\ 
Andreas Steiner \\ 
Lawrence Stewart \\ 
Ilya Tolstikhin \\ 
Michael Tschannen \\ 
Anton Tsitsulin \\ 
Nino Vieillard \\ 
Renjie Wu \\ 
Pingmei Xu \\ 
Haichuan Yang \\ 
Edouard Yvinec \\ 
Biao Zhang \\
Li Zhang \\ 
Joe Zou \\
\end{multicols}

\noindent\textbf{Contributors} \\

\begin{multicols}{3}
\noindent
Nicolas Aagnes \\ 
Abdelrahman Abdelhamed \\ 
Jakub Adamek \\
Shivani Agrawal \\ 
Shubham Agrawal \\ 
Ibrahim Alabdulmohsin \\ 
Jean Baptiste Alayrac \\ 
Uri Alon \\ 
Chandramouli Amarnath \\ 
Ankesh Anand \\ 
Chrysovalantis Anastasiou \\ 
Setareh Ariafar \\ 
François-Xavier Aubet \\ 
Kyriakos Axiotis \\ 
Federico Barbero \\ 
Joelle Barral \\ 
Alexei Bendebury \\ 
Urs Bergmann \\ 
Stanley Bileschi \\ 
Kat Black \\ 
Mathieu Blondel \\ 
Sebastian Borgeaud \\ 
Arthur Bražinskas \\ 
Ryan Burnell \\ 
Robert Busa-Fekete \\ 
Mu Cai \\ 
Daniele Calandriello \\
Glenn Cameron \\ 
Charlotte Caucheteux \\ 
Rahma Chaabouni \\ 
Garima Chadha \\ 
Jetha Chan \\ 
Blake Jianhang Chen \\ 
Jesse Chen \\ 
Lin Chen \\ 
Xu Chen \\ 
Derek Cheng \\ 
Tzu-hsiang Chien \\ 
Nikolai Chinaev \\ 
Yi Chou \\ 
Zhaohui Chu \\ 
Benjamin Coleman \\ 
Pooja Consul \\ 
Sam Conway-Rahman \\ 
Scott Crowell \\ 
Dylan Cutler \\ 
Vivek Dani \\ 
Samira Daruki \\ 
Anil Das \\ 
Daniel Deutsch \\ 
Nishanth Dikkala \\ 
Li Ding \\ 
Qiuhan Ding \\ 
Shenil Dodhia \\ 
Konstantin Donhauser \\ 
Tulsee Doshi \\ 
Anca Dragan \\ 
Alex Druinsky \\ 
Sahil Dua \\ 
Zoltan Egyed \\ 
Danielle Eisenbud \\ 
Daniel Eppens \\ 
Cindy Fan \\ 
Bahare Fatemi \\ 
Yassir Fathullah \\ 
Vlad Feinberg \\ 
Milen Ferev \\ 
Sebastian Flennerhag \\ 
Takumi Fujimoto \\ 
João Gabriel Oliveira \\ 
Isaac Galatzer-Levy \\ 
João Gante \\ 
Simon Geisler \\ 
Soham Ghosal \\ 
Antonious M. Girgis \\ 
Tamara von Glehn \\ 
Alec Go \\ 
Alhaad Gokhale \\ 
Alex Grills \\ 
Yiming Gu \\ 
Mayank Gupta \\
Pramod Gupta \\ 
Guru Guruganesh \\ 
Raia Hadsell \\ 
Hamza Harkous \\ 
Jitendra Harlalka \\ 
Demis Hassabis \\ 
Anja Hauth \\ 
Joe Heyward \\ 
Arian Hosseini \\ 
Chih-Yang Hsia \\ 
I-Hung Hsu \\ 
Xiaopeng Huang \\ 
Yangsibo Huang \\ 
Kevin Hui \\ 
Adrian Hutter \\ 
Te I \\ 
Fotis Iliopoulos \\ 
Advait Jain \\ 
Ganesh Jawahar \\ 
Ziwei Ji \\ 
Qilin Jin \\ 
Melvin Johnson \\ 
Kandarp Joshi \\ 
Arun Kandoor \\ 
Wang-Cheng Kang \\ 
Koray Kavukcuoglu \\ 
Mehran Kazemi \\ 
Kathleen Kenealy \\ 
Amr Khalifa \\ 
Phoebe Kirk \\ 
Ivan Korotkov \\ 
Suraj Kothawade \\ 
Vitaly Kovalev \\ 
Neel Kovelamudi \\ 
Adam Kraft \\ 
Ravin Kumar \\ 
Vivek Kumar \\
Harish Kuppam \\ 
Justin Lannin \\ 
Chen-Yu Lee \\ 
Seungji Lee \\ 
Dmitry Lepikhin \\ 
Alon Levkovitch \\ 
Dongdong Li \\ 
Qiujia Li \\ 
Valentin Liévin \\ 
Ethan Lin \\ 
Ziqian Lin \\ 
Casper Liu \\ 
Tianlin Liu \\ 
Tianqi Liu \\ 
Xin Liu \\ 
Ivan Lobov \\
Mayank Lunayach \\ 
Min Ma \\ 
Gagan Madan \\ 
Andrii Maksai \\ 
Eric Malmi \\ 
Michal Matuszak \\ 
Daniel McDuff \\ 
Gaurav Menghani \\ 
Maciej Mikuła \\ 
Daniil Mirylenka \\ 
Karolis Misiunas \\ 
Vedant Misra \\ 
Andreea Mitran \\ 
Kareem Mohamed \\ 
Maksim Mukha \\ 
Eric Noland \\ 
James O'Donnell \\ 
Brendan O'Donoghue \\ 
Kate Olszewska \\ 
Bernett Orlando \\ 
Wanqiong Pan \\ 
Rina Panigrahy \\ 
Unnati Parekh \\ 
Nicolas Perez-Nieves \\ 
Chunjong Park \\ 
Eric Paskie \\ 
Liqian Peng \\ 
Bryce Petrini \\ 
Slav Petrov \\ 
Jonas Pfeiffer \\ 
Bilal Piot \\ 
Martyna Plomecka \\ 
Siim Poder \\ 
Octavio Ponce \\ 
Arijit Pramanik \\ 
David Racz \\ 
Anish Rajan \\ 
Michelle Ramanovich \\ 
Anand Rao \\ 
Marvin Ritter \\ 
Vitor Rodrigues \\ 
Evan Rosen \\ 
Mikołaj Rybiński \\ 
Noveen Sachdeva \\ 
Michaël E. Sander \\  
Rohit Sathyanarayana \\ 
Sagar Savla \\ 
Samuel Schmidgall \\ 
Tal Schuster \\ 
George Scrivener \\ 
Benoit Seguin \\ 
Andrew Sellergren \\ 
Aliaksei Severyn \\ 
Izhak Shafran \\ 
Dhruv Shah \\ 
Bobak Shahriari \\ 
Yuan Shangguan \\ 
Ashish Shenoy \\ 
Pradeep Shenoy \\ 
Rakesh Shivanna \\ 
Pauline Sho \\ 
Lucas Spangher \\ 
Wojciech Stokowiec \\ 
Tim Strother \\ 
Yao Su \\ 
Yinghao Sun \\ 
Mukund Sundararajan \\ 
Andrea Tacchetti \\ 
Mor Hazan Taege \\ 
Pouya Tafti \\ 
Jean Tarbouriech \\ 
Chetan Tekur \\ 
Shantanu Thakoor \\ 
Rahul Thapa \\ 
Madeleine Traverse \\ 
Lenart Treven \\ 
Tao Tu \\ 
Chien Te Tung  \\ 
Çağlar Ünlü \\ 
Petar Veličković \\ 
Malini Pooni Venkat \\ 
Sagar Gubbi Venkatesh \\ 
Vidya Venkiteswaran \\ 
Francesco Visin \\ 
Alex Vitvitskyi \\ 
Kiran Vodrahalli \\ 
Weiyi Wang \\ 
Xin Wang \\ 
Tris Warkentin \\ 
Jan Wassenberg \\ 
John Wieting \\ 
Cindy Wu \\
Lechao Xiao \\ 
Hao Xu \\ 
Yuhui Xu \\ 
Fuzhao Xue \\
Arun Yadav \\ 
Jun Yan \\ 
Antoine Yang \\ 
Lin Yang \\ 
Ming-Hsuan Yang \\ 
Ziyu Ying \\ 
Jae Hyeon Yoo \\
Morteza Zadimoghaddam \\
Sajjad Zafar \\ 
Fred Zhang \\ 
Jiageng Zhang \\ 
Jianyi Zhang \\ 
Xiaofan Zhang \\ 
Chao Zhao \\ 
David Zhou \\ 
Chen Zou \\ 
\end{multicols}

\onecolumn
\section*{Appendix}

\noindent \textbf{Conversation format.} We give an example of a conversation including thinking, function definition and function calling in Table~\ref{tab:formatting_tokens}.

\begin{figure}[ht!]
    \centering
    \includegraphics[width=0.5\linewidth]{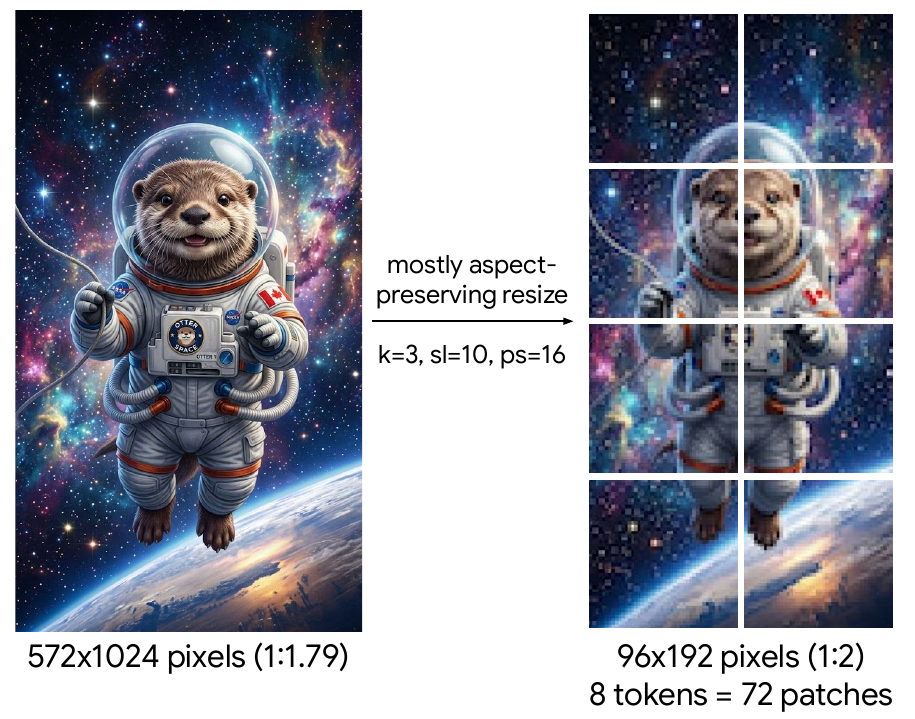}
    \caption{
    Image resizing. Here we use
    \texttt{patch\_size=16},
    \texttt{pooling\_kernel\_size=3},
    \texttt{max\_soft\_tokens=10}.
    The image is thus first resized to 2 $\times$ 4 pooled patches (each of size $48\text{px} \times 48\text{px}$), which is the closest match that results in a sequence length below the targeted 10. The 72 patches (each of size $16\text{px} \times 16\text{px}$) are then processed by the vision encoder, the vision encoder representations are pooled $3\times3$, and the resulting 8 soft tokens are processed by the LLM backbone.
    }
    \label{fig:varres}
\end{figure}

\noindent \textbf{Vision.} We detail the vision encoder architecture in Table~\ref{tab:vision_encoder_comparison}. We then illustrate how images are resized before being fed to the vision encoder in Figure~\ref{fig:varres}, and detail the resizing algorithm in Algorithm~\ref{alg:resize}. We display the vision benchmark scores of Gemma 4 models at low resolution ($N_{max} = 280$) in Table~\ref{tab:it_vis_sl_280}.

\begin{table}[H]
\centering
\captionsetup{justification=centering}
\begin{tabular}{@{} c c c c c @{}}
\toprule
\textbf{Total Params} & $d_{model}$ & $d_{MLP}$ & $N_{heads}$ & $N_{layers}$ \\
\midrule
\textbf{550M} & 1152 & 4304 & 16 & 27 \\
\textbf{150M} &  768 & 3072 & 12 & 16 \\
\bottomrule
\end{tabular}
\caption{Vision encoder architecture.}
\label{tab:vision_encoder_comparison}
\end{table}

\begin{algorithm}
  \caption{Aspect-Ratio Preserving Image Resizing (see also Figure~\ref{fig:varres})}
  \label{alg:resize}
  \begin{algorithmic}[1]
    \Require Image $\mathbf{I} \in \mathbb{R}^{H \times W \times C}$, patch size $p$, max tokens $N_{\max}$, pooling kernel size $k$
    \State $m \gets k \cdot p$ \Comment{Pooled patch size}
    \State $T \gets N_{\max} \cdot m^2$ %
    \State $f \gets \sqrt{T / (H \cdot W)}$ \Comment{Ideal scaling factor}
    \State $H_{\text{ideal}} \gets f \cdot H$
    \State $W_{\text{ideal}} \gets f \cdot W$
    \State $H_{\text{target}} \gets \lfloor H_{\text{ideal}} / m \rfloor \cdot m$ \Comment{Round down}
    \State $W_{\text{target}} \gets \lfloor W_{\text{ideal}} / m \rfloor \cdot m$
    \State $\mathbf{I}_{\text{resized}} \gets \text{BicubicResize}(\mathbf{I}, H_{\text{target}}, W_{\text{target}})$
    \State \textbf{return} $\mathbf{I}_{\text{resized}}$
  \end{algorithmic}
  \label{alg:resize}
\end{algorithm}

\begin{table*}[t]
    \centering
    \footnotesize
    \begin{tabular}{p{0.3\linewidth} p{0.4\linewidth}}
    \toprule
    \textbf{Context} & \textbf{Formatting} \\
        \midrule
        Thinking toggle & \texttt{\color{Plum}<|think|>} \\
        \midrule
        Function declaration & \texttt{\color{Orange}<|tool>declaration:...<tool|>} \\
        \midrule
        Function call & \texttt{\color{RedOrange}<|tool\_call>call:...<tool\_call|>} \\
        \midrule
        Thinking trace & \texttt{\color{SeaGreen}<|channel>thought \dots <channel|>} \\
        \midrule
        System turn & \texttt{\color{NavyBlue}<|turn>system} \\
        \midrule
        User turn & \texttt{\color{NavyBlue}<|turn>user} \\
        \midrule
        Model turn & \texttt{\color{NavyBlue}<|turn>model} \\
        \midrule
        End of turn & \texttt{\color{NavyBlue}<turn|>} \\
         \midrule
    \multicolumn{2}{c}{\textbf{Example of discussion:}} \\
    \midrule
   \multicolumn{2}{p{0.8\linewidth}}{
        \textbf{Toggle thinking mode.}\par
        \textbf{Declare function.}\par
        \textbf{User:} \texttt{I want you to book a train ticket for me.}\par
        \textbf{Model:} \texttt{<\ldots> Where would you like to go?}\par
        \textbf{User:} \texttt{To Rome.}\par
        \textbf{Model:} \texttt{<\ldots> Looking for available tickets: <function call>}
    } \\
    \midrule
    \multicolumn{2}{c}{\textbf{Model input:}} \\
    \midrule
   \multicolumn{2}{p{0.8\linewidth}}{
        \texttt{\color{red}[BOS]}\par
        \texttt{\color{NavyBlue}<|turn>system}\par
        \texttt{\color{Plum}<|think|>}\par
        \texttt{}\par
        \texttt{\texttt{\color{Orange}<|tool>declaration:search\_train\{\ldots\}<tool|>\color{NavyBlue}<turn|>}}\par
        \texttt{\color{NavyBlue}<|turn>user}\par
        \texttt{I want you to book a train ticket for me.}\texttt{\color{NavyBlue}<turn|>}\par
        \texttt{\color{NavyBlue}<|turn>model}\par
        \texttt{\color{SeaGreen}<|channel>thought \dots <channel|>\color{Black}Where would you like to go?\color{NavyBlue}<turn|>}\par
        \texttt{\color{NavyBlue}<|turn>user}\par
        \texttt{To Rome.}\texttt{\color{NavyBlue}<turn|>}\par
        \texttt{\color{NavyBlue}<|turn>model}
    } \\
    \midrule
    \multicolumn{2}{c}{\textbf{Model output:}} \\
    \midrule
   \multicolumn{2}{p{0.8\linewidth}}{
        \texttt{\color{SeaGreen}<|channel>thought \dots <channel|>\color{Black}Looking for available tickets:}\par 
        \texttt{\color{RedOrange}<|tool\_call>call:search\_train\{from:<|"|>Athens<|"|>,to:<|"|>Rome<|"|>\}}\par
        \texttt{\color{RedOrange}<tool\_call|>}\texttt{\color{NavyBlue}<turn|>}
    } \\
    \bottomrule
    \end{tabular}
    \caption{Formatting for Gemma IT models. Explicitly add the \texttt{\color{red}[BOS]} token after tokenization, or use the \texttt{add\_bos=True} option in the tokenizer. \textit{Do not tokenize the text "[BOS]"}. Add \texttt{\color{Plum}<|think|>} in a leading system turn to activate the thinking mode. Check the official documentation for the function declaration and function calling syntax, as well as more advanced examples.}
    \label{tab:formatting_tokens}
    \vspace{-0.5cm}
\end{table*}

\begin{table*}[th!]
{
\footnotesize
\centering
\setlength{\tabcolsep}{5pt} 
\begin{tabular}{@{}l ccccc c @{}}
\toprule
\multirow{2}{*}[-0.2em]{} & \multicolumn{5}{c}{Gemma~4} \\
\cmidrule{2-6}\cmidrule{7-7}
& 31B & \makecell{26B-A4B} & 12B & E4B & E2B \\
\midrule
MMMU Pro \nocite{yue2025mmmu} & 75.8 & 73.2 & 67.7 & 51.4 & 43.2 \\
MATH-Vision \nocite{wang2024measuring} & 83.4 & 80.3 & 76.7 & 59.2 & 53.0 \\
MedXPertQA MM \nocite{zuo2025medxpertqa} & 60.7 & 55.7 & 47.4 & 28.7 & 22.5 \\
InfographicVQA \nocite{mathew2022infographicvqa} & 82.8 & 77.8 & 58.7 & 54.8 & 44.6 \\
OmniDocBench 1.5 $\downarrow$ \nocite{ouyang2025omnidocbench} & 0.201 & 0.269 & 0.408 & 0.307 & 0.496 \\
\bottomrule
\end{tabular}
\caption{Gemma~4 models performance on vision benchmarks at resolution $N_{max}$ = 280 (thinking).}
\label{tab:it_vis_sl_280}
}
\end{table*}

\end{document}